\documentclass{article}

    \usepackage[preprint]{neurips_2025}

\usepackage[utf8]{inputenc} 
\usepackage[T1]{fontenc}    
\usepackage{hyperref}       
\usepackage{url}            
\usepackage{booktabs}       
\usepackage{amsfonts}       
\usepackage{nicefrac}       
\usepackage{microtype}      
\usepackage{xcolor}         

\usepackage{arydshln}
\usepackage{subcaption}
\usepackage{multicol}
\usepackage{multirow}
\usepackage{enumitem} 
\setlist{nosep, leftmargin=*, align=left}

\usepackage{amsmath}
\usepackage{amssymb}
\usepackage{mathtools}
\usepackage{amsthm}
\DeclareMathOperator{\rank}{rank}

\DeclareMathOperator*{\argmin}{arg\,min}
\theoremstyle{plain}
\newtheorem{theorem}{Theorem}[section]

\newtheorem{corollary}[theorem]{Corollary}
\theoremstyle{definition}
\newtheorem{definition}[theorem]{Definition}

\theoremstyle{remark}
\newtheorem{remark}[theorem]{Remark}
\usepackage{enumitem}

\usepackage[capitalize,noabbrev]{cleveref}

\title{Towards Achieving Perfect Multimodal Alignment}

\author{
Abhi Kamboj$^1$ \qquad \qquad \qquad Minh N. Do$^1$ \\ \\
University of Illinois at Urbana-Champaign$^1$ \\
{\tt\small \{akamboj2, minhdo\}@illinois.edu}
}

\begin{document}

\maketitle

\begin{abstract}
Multimodal alignment constructs a joint latent vector space where modalities representing the same concept map to neighboring latent vectors. 
We formulate this as an inverse problem and show that, under certain conditions, paired data from each modality can map to equivalent latent vectors, which we refer to as perfect alignment.
When perfect alignment cannot be achieved, it can be approximated using the Singular Value Decomposition (SVD) of a multimodal data matrix.
Experiments on synthetic multimodal Gaussian data verify the effectiveness of our perfect alignment method compared to a learned contrastive alignment method.
We further scale to a practical application of cross-modal transfer for human action recognition and show that perfect alignment boosts the accuracy of the model. 
We conclude by discussing how these findings can be applied to various modalities and tasks and the limitations of our method.
We hope these findings inspire further exploration of perfect alignment and its applications in representation learning.
\end{abstract}


\section{Introduction}
\label{sec:intro}
Humans naturally perceive the same concept through multiple senses, a capability artificial intelligence (AI) aims to replicate with multimodal data. 
However, integrating diverse data types remains challenging due to differences in abundance, information richness, and annotation difficulty. For example, images and videos are plentiful and often easy to label, while modalities like MRI, ECG, or IMU are scarce and harder to annotate. 
This diversity raises a fundamental question: How can we establish a unified representational framework that enables effective knowledge transfer across modalities, particularly from data-rich sources to more specialized, resource-constrained domains?

To interpret multimodal data, AI methods typically align the semantic meanings of different modalities within a shared latent space at the output of modality-specific encoders.
For instance, models can associate an image with descriptive text~\cite{radford2021learning} or with corresponding sounds, videos or other sensors~\cite{Girdhar_2023_CVPR}. 
While such alignment is often achieved through large-scale learned methods and specialized loss functions, these approaches remain fundamentally approximate and often lack theoretical rigor and interpretability.

\begin{figure*}[t]
\centering
\includegraphics[width=\linewidth]{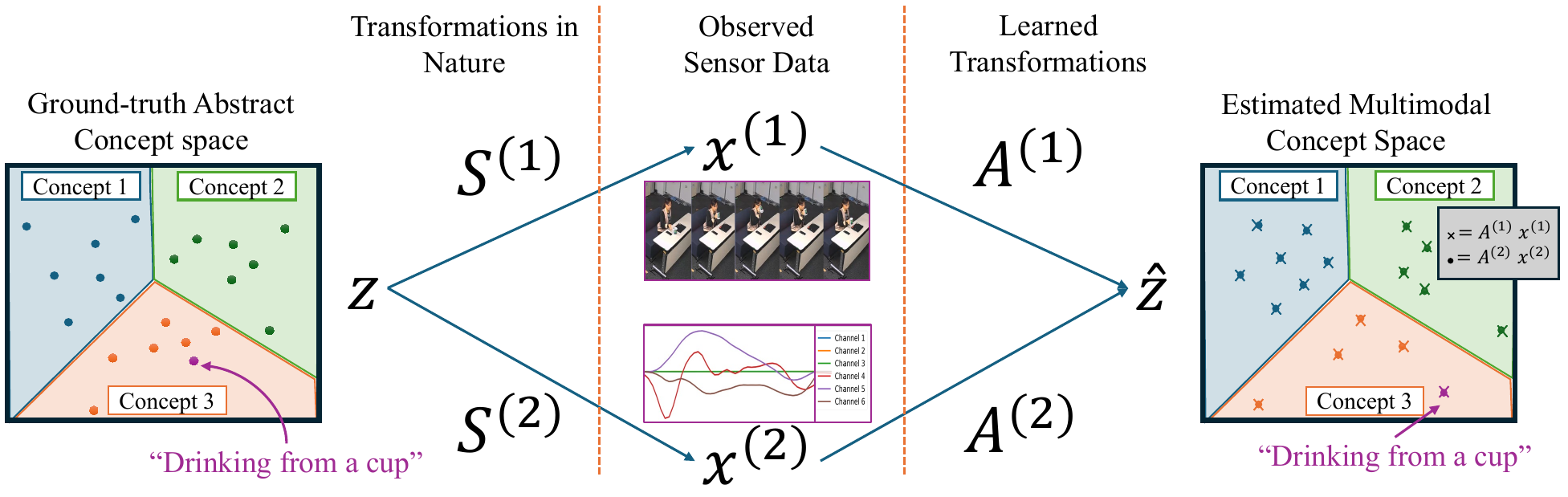}
\caption{\textbf{Data generation model:} Latent concepts $\mathbf{z}_i$ are transformed through modality-specific matrices $\mathbf{S}^{(1)}, \mathbf{S}^{(2)}$ to generate observations in different modalities $\mathbf{x}_i^{(1)}, \mathbf{x}_i^{(2)}$. Our goal is to recover alignment matrices $\mathbf{A}^{(1)}, \mathbf{A}^{(2)}$ that invert these transformations. Ideally, the learned transformation preserves the latent structure such that a sample from a concept class in the original latent space remains in that class grouping in the estimated latent space.}
\label{fig:data_gen}
\end{figure*}

Prior work explores contrastive alignment through geometric~\cite{wang2020understanding}, probabilistic~\cite{chen2024your,che2025law}, and information-theoretic~\cite{oord2018representation,poole2019variational} perspectives. 
However, these analyses primarily reinterpret existing alignment frameworks rather than proposing new methodologies. 
Following the Platonic Representation Hypothesis~\cite{huh2024platonic}, we model the existence of a ground-truth representation space of semantic concepts from which modalities are generated through transformations in nature. 
Thus, our learned alignment transformations must invert these natural transformations, as depicted in \cref{fig:data_gen}.
In this work, we reframe multimodal alignment as a linear inverse problem, a class of problems well-studied in linear algebra and signal processing~\cite{harikumar1999perfect}.
This reframing enables the derivation of a representation space with perfect alignment between two modalities. 
We define perfect alignment as the existence of modality-specific encoders that map training instances from distinct modalities to identical latent representations.
Notably, we demonstrate that empirical risk minimization and linear regression emerge as special cases of this framework, bridging classical machine learning paradigms to multimodal alignment.

This work proves that if a perfect alignment mapping exists, the transformations can be recovered by extracting the last few left singular vectors of a multimodal data matrix where one modality is additively inverted.
The formulation extracts the best approximation when perfect alignment cannot be achieved.
Despite limitations regarding assumptions on data dimensions, latent size, and number of data points, this methodology provides a framework for using linear algebra intuition to understand conditions under which alignment succeeds or fails.
Additionally, Perfect Alignment demonstrates strong connections to linear regression and statistical learning theory, establishing pathways for extending results from these domains to alignment problems.

The perfect alignment approach is first validated on low-dimensional synthetic multimodal data, illustrating the method's effectiveness in producing perfect alignment.
Notably, although the exact synthetic ground-truth latent space cannot be recovered, the perfect alignment method preserves the general distribution shape of the original latent space.
This preservation implies that decision boundaries learned in the true latent concept space can also be learned in the estimated space.

The method is subsequently evaluated on a real-world dataset of human actions.
Given the assumption of linear transformations, Perfect Alignment is hypothesized to be effective when leveraged on top of a pretrained encoder.
A method for cross-modal transfer from videos to inertial sensors is developed using contrastive learning.
We then demonstrate that adding a perfect alignment step on top of the pretrained encoders improves the transfer accuracy of the model.
Visualization of learned representations before and after alignment reveals that post-alignment, modalities and their corresponding pairs overlap with greater frequency.
The contributions are:
\begin{itemize}
\item \textbf{Theoretical Framework:} A novel inverse problem formulation for multimodal perfect alignment, a closed-form solution for perfect alignment under certain conditions, and an analysis of this framework and its connection to existing machine learning theory.
\item \textbf{Empirical Validation:} Successful alignment on synthetic data, experiments that show this alignment follows mathematical intuition, and a real-world example of how perfect alignment can boost cross-modal transfer performance for human action recognition.
\end{itemize}

\section{Perfect Alignment Formulation}
\label{sec:methods}

\label{sec:methods:perfect_alignment}



We propose a method to achieve \textbf{perfect alignment} between two modalities by solving an inverse problem to construct the aligned latent space. To formalize this, we first define key notation and assumptions.  
\begin{itemize}[leftmargin=*,noitemsep]
    \item Superscripts (e.g., $^{(1)}$, $^{(2)}$) denote modalities. 
    \item Subscripts (e.g., $_i$, $_j$) index samples in a dataset.  
    \item Lowercase bold letters (e.g., $\mathbf{x}$, $\mathbf{z}$) represent vectors.  
    \item Uppercase bold letters (e.g., $\mathbf{A}$, $\mathbf{X}$) indicate matrices.  
    \item Calligraphic letters (e.g., $\mathcal{X}$, $\mathcal{Z}$) denote vector spaces or sets.  
\end{itemize}

\begin{definition}[Perfect Alignment]
\label{def:perfect_alignment}
Let $\mathcal{X}^{(1)}$ and $\mathcal{X}^{(2)}$ denote the input spaces of two modalities, with $\mathcal{Z}$ being their shared latent space. Given a dataset $\mathcal{D} = \{(\mathbf{x}_i^{(1)}, \mathbf{x}_i^{(2)})\}_{i=1}^n$ of corresponding multimodal instances, \emph{perfect alignment} is defined by the existence of encoder functions $f^{(1)}: \mathcal{X}^{(1)} \to \mathcal{Z}$ and $f^{(2)}: \mathcal{X}^{(2)} \to \mathcal{Z}$ satisfying:
\begin{equation}
\forall (\mathbf{x}^{(1)}, \mathbf{x}^{(2)}) \in \mathcal{D},\quad f^{(1)}(\mathbf{x}^{(1)}) = f^{(2)}(\mathbf{x}^{(2)}) = \mathbf{z},
\end{equation}
where $\mathbf{z} \in \mathcal{Z}$ is the unified semantic representation of the shared concept underlying the pair $(\mathbf{x}^{(1)}, \mathbf{x}^{(2)})$.
\end{definition}



Typical representation learning methods start with observed sensor data $\mathbf{x}$, and attempt to learn transformations to some feature space $\mathbf{z}$.
For multimodal alignment, methods often aim  to learn transformations from the same concept represented through different modalities, to feature representations that are correlated, or are the same (as is the case in Perfect Alignment). 
To better understand and formalize this process, let's posit that this there exists some groundtruth feature space that our methods are trying to recover, and some transformations in nature that generate our sensor data, which our learned transformations are trying to invert. 
\cref{fig:data_gen} depicts this model.

Let $\mathcal{Z} \subseteq \mathbb{R}^k$ be a ground truth latent space representing semantic concepts.
We assume each modality $m \in {1,2}$ is generated via linear transformations:
\begin{equation}
\mathbf{x}_i^{(m)} = \mathbf{S}^{(m)} \mathbf{z}_i,
\end{equation}
where for sample $i$:
\vspace{-.5em}
\begin{itemize}[leftmargin=*,noitemsep]
\item $\mathbf{z}_i \in \mathcal{Z} \subseteq \mathbb{R}^k$ is the latent concept vector
\item $\mathbf{S}^{(m)} \in \mathbb{R}^{d_m \times k}$ is the modality-specific generation matrix
\item $\mathbf{x}_i^{(m)} \in \mathcal{X}^{(m)} \subseteq \mathbb{R}^{d_m}$ is the observed data in modality $m$
\end{itemize}

For aligned pairs $(\mathbf{x}_i^{(1)}, \mathbf{x}_i^{(2)})$ generated from the same $\mathbf{z}_i$, our goal is to recover projection matrices $\mathbf{A}^{(1)} \in \mathbb{R}^{k \times d_1}$ and $\mathbf{A}^{(2)} \in \mathbb{R}^{k \times d_2}$ such that:
\begin{equation}
\mathbf{A}^{(1)} \mathbf{x}_i^{(1)} = \mathbf{A}^{(2)} \mathbf{x}_i^{(2)} = \mathbf{z}_i \quad \forall i \in {1,\dots,n}.
\end{equation}

Assuming our model does not have access to the ground truth latent vector $\mathbf{z}_i$, this reduces to solving the system:
\begin{equation}
\mathbf{A}^{(1)} \mathbf{X}^{(1)} - \mathbf{A}^{(2)} \mathbf{X}^{(2)} = 0,
\label{eq:alignment_system}
\end{equation}
where $\mathbf{X}^{(m)} = [\mathbf{x}_1^{(m)} \cdots \mathbf{x}_n^{(m)}] \in \mathbb{R}^{d_m \times n}$ in which each column $\mathbf{x}_i^{(m)}$ represents the $i$-th data point in modality $m$.





When \cref{eq:alignment_system} holds, we achieve \textbf{perfect alignment} as defined in \cref{def:perfect_alignment}, where the encoder functions $f^{(1)}$ and $f^{(2)}$ correspond to the linear transformations $\mathbf{A}^{(1)}$ and $\mathbf{A}^{(2)}$, respectively.
To determine these matrices we construct the combined matrices:
\setlength{\dashlinedash}{1pt}
\setlength{\dashlinegap}{1pt}
\setlength{\arrayrulewidth}{0.5pt}
\begin{equation}
\small
    \mathbf{A} = \left[\begin{array}{c:c} \mathbf{A}^{(1)} & \mathbf{A}^{(2)} \end{array}\right] \in \mathbb{R}^{k \times d}, \quad 
    \mathbf{X} = \left[ \begin{array}{c} \mathbf{X}^{(1)} \\ \hdashline -\mathbf{X}^{(2)} \end{array}\right] \in \mathbb{R}^{d \times n},
    \label{eq:stacked_matrices}
\end{equation}
where $d = d_1 + d_2$.

This allows us to rewrite \cref{eq:alignment_system} as the linear inverse problem:
\begin{equation}
\mathbf{A}\mathbf{X} = \mathbf{0},
\label{eq:perf_align}
\end{equation}
where $\mathbf{0} \in \mathbb{R}^{k \times n}$ is the zero matrix. The goal is to find a non-trivial solution $\mathbf{A} \neq \mathbf{0}$ that satisfies this equation.





\begin{theorem}[Existence of Perfect Alignment]
\label{thm:perfect_alignment}
Given the inverse problem $\mathbf{A}\mathbf{X} = \mathbf{0}$ defined in \cref{eq:perf_align}, where $\mathbf{X} \in \mathbb{R}^{d \times n}$ is a given data matrix and $\mathbf{A} \in \mathbb{R}^{k \times d}$ is unknown, if $\mathbf{X}$ has a left null space $\mathcal{N}(\mathbf{X}^T)$ of dimension $\dim(\mathcal{N}(\mathbf{X}^T)) \geq k$, then there exists a closed-form solution for $\mathbf{A}$. Specifically, the rows of $\mathbf{A}$ can be formed by any $k$ linearly independent vectors spanning $\mathcal{N}(\mathbf{X}^T)$.
\end{theorem}

\begin{proof}
The proof involves recognizing that any vector $\mathbf{a}$ in the left null space of $\mathbf{X}$ satisfies $\mathbf{a}^T\mathbf{X} = 0$. Therefore, if $\mathbf{X}$ has a null space of dimension at least $k$, we can select $k$ linearly independent vectors from this null space to form the rows of $\mathbf{A}$. This ensures that $\mathbf{A}\mathbf{X} = 0$ is satisfied. 
A full proof is given in Appendix \cref{app:proof_perfect_alignment}.
\end{proof}

\begin{corollary}[Approximate Alignment]
\label{cor:approximate_alignment}
If $\mathbf{X} \in \mathbb{R}^{d \times n}$ has a left null space $\mathcal{N}(\mathbf{X}^T)$ with $\dim(\mathcal{N}(\mathbf{X}^T)) < k$, an approximation to $\mathbf{A}\mathbf{X} = \mathbf{0}$ can be obtained by selecting the $k$ basis vectors corresponding to the smallest singular values of $\mathbf{X}$. This approximation minimizes the Frobenius norm $\|\mathbf{A}\mathbf{X}\|_F$.
\end{corollary}
\begin{proof}
    This is a direct application of Eckhart-Young-Mirsky theorem. The full proof is shown in \cref{app:proof_perfect_alignment}.
\end{proof}





\paragraph{Method for Finding $\mathbf{A}$.}
To determine $\mathbf{A}$, compute the Singular Value Decomposition (SVD) of $\mathbf{X}$:
\[
\mathbf{X} = \mathbf{U} \mathbf{\Sigma} \mathbf{V}^T,
\]
where $\mathbf{U} \in \mathbb{R}^{d \times d}$. 
Assuming $k\leq d$, Extract the last $k$ columns of $\mathbf{U}$, denoted $\mathbf{u}_{d-k+1}, \dots, \mathbf{u}_d$, which correspond to the basis vectors of the left null space of $\mathbf{X}$ (if $\dim(\mathcal{N}(\mathbf{X}^T)) \geq k$) or its smallest singular values (otherwise). The solution for $\mathbf{A}$ is:
\begin{equation}    
    \mathbf{A}^* = \begin{bmatrix} \mathbf{u}_{d-k+1}^T \\ \vdots \\ \mathbf{u}_d^T \end{bmatrix},
    \label{eq:align_soln}
\end{equation}
where $\mathbf{u}_j$ is the $j^\text{th}$ column of $\mathbf{U}$. This method achieves perfect alignment when $\dim(\mathcal{N}(\mathbf{X}^T)) \geq k$ and an optimal approximation otherwise.

\begin{remark}[Assumption on $k$ and $d$]
\label{remark:kd}
The assumption $k \leq d$ is valid because the latent space dimension $k$ is typically smaller than the data dimension $d$ in representation learning. This reflects the common goal of compressing high-dimensional data into a lower-dimensional space while preserving essential semantic information.
\end{remark}

\begin{remark}[Achievability and Computational Cost of Perfect Alignment]
Perfect alignment is often not achievable in practice because the number of data points $n$ is large, thus $\mathbf{X}$ becomes a wide matrix and the left null space of $\mathbf{X}$ has limited dimensionality. Furthermore, computing $\mathbf{A}$ via full SVD of $\mathbf{X} \in \mathbb{R}^{d \times n}$ has a time complexity of $O(d^2n + dn^2 + n^3)$. For large $d$ and $n$, this becomes prohibitive, motivating approximate methods like gradient descent. When only the $k$ smallest singular vectors are needed, truncated SVD reduces this to $O(dnk)$, making it feasible for moderate $k$.
\end{remark}

\begin{remark}[Comparison to Linear Regression]
Standard linear regression minimizes $\argmin_\mathbf{W} \|\mathbf{Y} - \mathbf{W}\mathbf{X}\|_F^2$, which is a special case of our alignment objective $\argmin_\mathbf{A} \|\mathbf{A}\mathbf{X}\|_F^2$ when:
\begin{itemize}[leftmargin=*,noitemsep]
    \item $\mathbf{A}^{(1)} = \mathbf{I}_d$ (identity mapping for modality 1)
    \item $\mathbf{X}^{(1)} = \mathbf{Y}$ (one modality is the regression target)
    \item $\mathbf{A}^{(2)} = \mathbf{W}$ (learned regression weights for modality 2)
    \item $\mathbf{X}^{(2)} = \mathbf{X}$ (second modality is the data)
\end{itemize}
\end{remark}

\begin{remark}[Connection to Empirical Risk Minimization (ERM)]
ERM seeks a model $f \in \mathcal{F}$ that minimizes the empirical risk:
\[
R_{\text{emp}}(f) = \frac{1}{n} \sum_{i=1}^n \ell(f(\mathbf{x}_i), y_i).
\]
Our framework extends this to two hypothesis classes $\mathcal{F}^{(1)}, \mathcal{F}^{(2)}$, minimizing:
\[
R_{\text{emp}}(f^{(1)}, f^{(2)}) = \frac{1}{n} \sum_{i=1}^n \ell\left(f^{(1)}(\mathbf{x}_i^{(1)}), f^{(2)}(\mathbf{x}_i^{(2)})\right),
\]
where $f^{(1)}, f^{(2)}$ are linear transformations $\mathbf{A}^{(1)}, \mathbf{A}^{(2)}$, and $\ell$ is the alignment loss $\|\mathbf{A}\mathbf{X}\|_F^2$.

\end{remark}




\subsection{Error Metrics for Perfect Alignment}
\label{sec:methods:error_metrics}

Let $\hat{\mathbf{z}}_i^{(m)} = \mathbf{A}^{(m)}\mathbf{x}_i^{(m)}$ denote the latent vectors estimated from modality $m$. We define two error metrics to evaluate alignment quality:

\begin{enumerate}[leftmargin=*,noitemsep]
    \item \textbf{Cross-Modal Alignment Error (CMAE)}: Quantifies the discrepancy between latent representations from different modalities. CMAE is computed as:
    \begin{equation}
        \text{CMAE} = \frac{1}{n} \sum_{i=1}^n \left\| \hat{\mathbf{z}}_i^{(1)} - \hat{\mathbf{z}}_i^{(2)} \right\|_2.
    \end{equation}
    Note that for normalized latent vectors (i.e., $\|\hat{\mathbf{z}}_i^{(m)}\|_2 = 1$), minimizing CMAE is equivalent to maximizing their cosine similarity—the same objective as the InfoNCE loss used in contrastive learning~\cite{oord2018representation}. \textit{Unlike contrastive methods, our framework does not assume normalized latent vectors; thus, cosine similarity is not directly applicable as a metric.}
    
    \item \textbf{Modality Latent Reconstruction Error (MLRE)}: Measures fidelity to the true latent space $\mathcal{Z}$, applicable only in synthetic experiments where $\mathbf{z}_i$ is known. For modality $m$, MLRE is:
    \begin{equation}
        \text{MLRE}^{(m)} = \frac{1}{n} \sum_{i=1}^n \left\| \mathbf{z}_i - \hat{\mathbf{z}}_i^{(m)} \right\|_2.
    \end{equation} 
     $\text{MLRE}=\sum_1^M\frac{1}{M}\text{MLRE}^{(m)}$ refers to the average for $M$ modalities.
\end{enumerate}

\section{Experiments}
\label{sec:experiments}

\subsection{Synthetic Data}
\label{sec:experiments:synthetic}
\paragraph{Data Generation.}
We generate synthetic data from a ground-truth latent space $\mathcal{Z} \subseteq \mathbb{R}^2$, modeled as a mixture of two Gaussian distributions:
\begin{equation}
    \mathcal{Z} \sim \pi_1 \mathcal{N}(\boldsymbol{\mu}_1, \boldsymbol{\Sigma}_1) + \pi_2 \mathcal{N}(\boldsymbol{\mu}_2, \boldsymbol{\Sigma}_2),
\end{equation}
where $\pi_1 = \pi_2 = 0.5$, $\boldsymbol{\mu}_1 = [0, 1]^T$, $\boldsymbol{\mu}_2 = [4, 5]^T$, and $\boldsymbol{\Sigma}_1 = \boldsymbol{\Sigma}_2 = \mathbf{I}_2$ (the 2D identity matrix). 

The data generation process proceeds as follows:
\begin{enumerate}[leftmargin=*,noitemsep]
    \item Sample $n = 2000$ vectors: $\mathcal{D}_Z = \{\mathbf{z}_i\}_{i=1}^{2000},  \mathbf{z}_i\sim \mathcal{Z}$.
    \item Project $\mathcal{D}_Z$ into two modalities using randomly generated matrices $\mathbf{S}^{(1)}, \mathbf{S}^{(2)} \in \mathbb{R}^{2 \times 2}$, where each entry is uniformly sampled from $[-5, 5]$:
    \begin{align}
        \mathcal{D}_{X} &= \{(\mathbf{x}^{(1)},\mathbf{x}^{(2)})|\mathbf{x}_i^{(1)} = \mathbf{S}^{(1)}\mathbf{z}_i, \mathbf{x}_i^{(2)} = \mathbf{S}^{(2)}\mathbf{z}_i\}_{i=1}^{2000}, \label{eq:mod_gen} 
    \end{align}
\end{enumerate}

\cref{fig:data_gen} illustrates this pipeline, showing the latent space clusters and their projections into the two modalities.

\paragraph{Alignment and Reconstruction Errors.}
\begin{figure}[t]
\begin{minipage}{.5\linewidth}
    \centering
    \includegraphics[width=\linewidth]{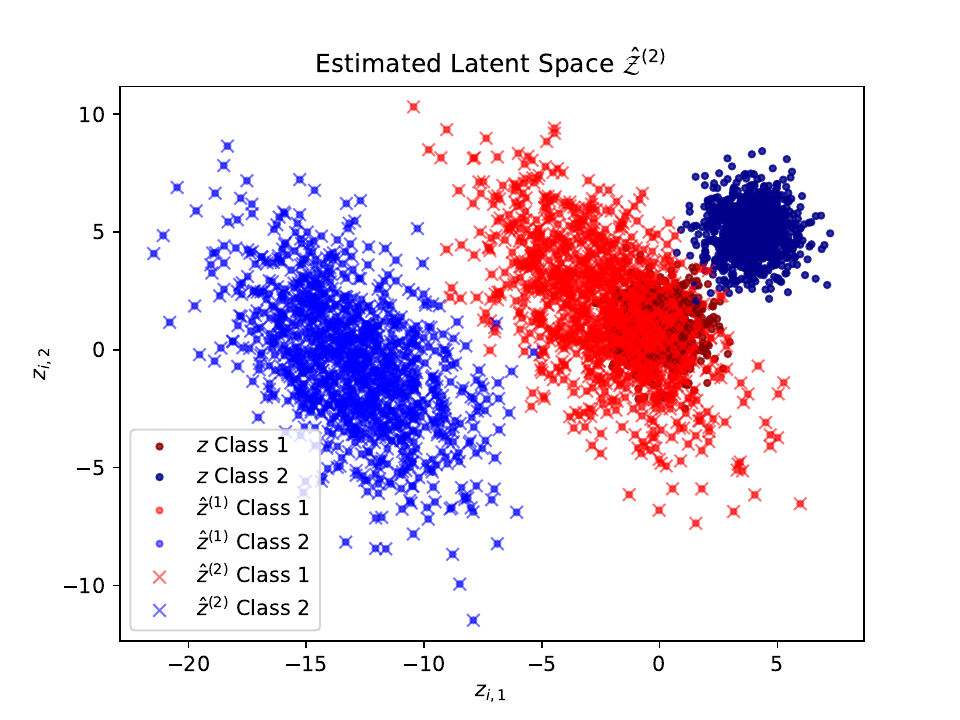}
    \caption{\textbf{Aligned Latent Space:} Recovered latent space $\hat{\mathcal{Z}}$ from synthetic data using our alignment method (see \cref{sec:experiments}). Colors denote ground-truth cluster membership.}
    \label{fig:z_hat_gmm}
\end{minipage}
\hfill
\begin{minipage}{.5\linewidth}
    \centering
    \caption{\textbf{Alignment Errors:} Modality Reconstruction Error (MRE), Cross-Modal Alignment Error (CMAE) and Normalized CMAE (NCMAE) across alignment methods. \textbf{Perfect Alignment} achieves near-zero CMAE (even after normalization), while contrastive alignment exhibits higher errors. Normalized CMAE computes alignment error for L2-normalized latent vectors.$\dagger$ \cite{nakada2023understanding}} 
    \label{tab:synth_errors}
    \resizebox{\linewidth}{!}{
    \begin{tabular}{lcccc}
    \toprule
    \textbf{Alignment} & $\text{MLRE}^{(1)}$ & $\text{MLRE}^{(2)}$ & \textbf{CMAE} & \textbf{NCMAE} \\
    \midrule
    Perfect (Ours) & 10.9 & 10.9 & $3.66e^{-15}$ & $3.14e^{-16}$ \\
    Contrastive$\dagger$ & 8.73 & 4.34 & 5.44 & 0.0298 \\
    \bottomrule
    \end{tabular}
    }
\end{minipage}
\end{figure}

Using the method in \cref{sec:methods:perfect_alignment}, we compute $\mathbf{A}^{(1)}$ and $\mathbf{A}^{(2)}$, then evaluate CMAE and MRE as defined in \cref{sec:methods:error_metrics}. \cref{tab:synth_errors} shows:

\begin{itemize}[leftmargin=*,noitemsep]
\item \textbf{Near-perfect alignment}: CMAE $\approx 3.66 \times 10^{-15}$ confirms latent representations from both modalities coincide almost exactly. This value is likely floating-point error.
\item \textbf{High reconstruction error}: MLRE $\approx 10.9$ indicates the estimated latent space $\hat{\mathcal{Z}}$ differs from the ground-truth $\mathcal{Z}$.
\end{itemize}

\paragraph{Key Insights.}  \Cref{fig:z_hat_gmm} visualizes $\hat{\mathcal{Z}}$, where aligned points $\hat{\mathbf{z}}_i^{(1)}$ and $\hat{\mathbf{z}}_i^{(2)}$ overlap perfectly but form clusters distinct from the original GMM. This arises because solutions to $\mathbf{A}\mathbf{X} = \mathbf{0}$ are non-unique—any linear transformation of the basis in \cref{eq:align_soln} yields valid solutions. While perfect alignment is achieved, perfect reconstruction requires identifying a specific transformation that maps $\hat{\mathcal{Z}}$ to $\mathcal{Z}$, a more constrained problem that requires more information about $\mathcal{Z}$
Despite high MLRE, the transformations preserve cluster structure (Gaussianity). This enables class separation in $\hat{\mathcal{Z}}$ and demonstrates that perfect alignment—not exact latent space recovery-may suffice for cross-modal transfer tasks.

\begin{figure}
\centering
\includegraphics[width=\linewidth]{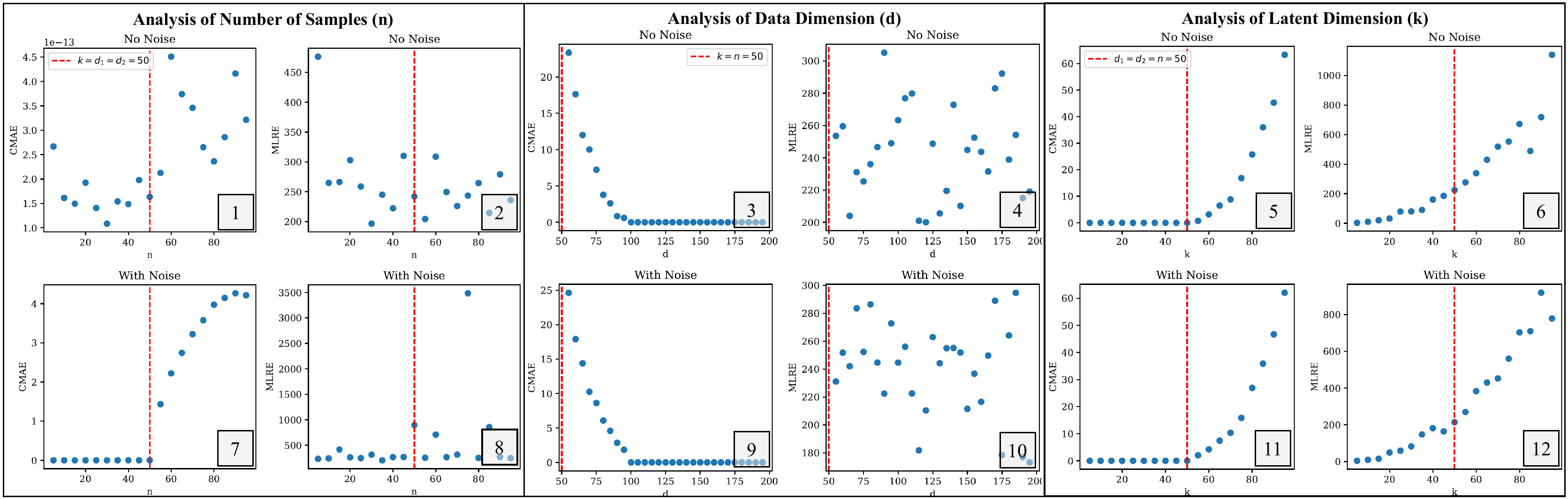}
\caption{\textbf{Robustness Analysis:} CMAE/MLRE results using the proposed perfect alignment solver under varying parameters.}
\label{fig:vary_params_noise}
\end{figure}

\paragraph{Parameter and Noise Experiments.}
We analyze the effect of $n$, $d$, and $k$ on the MRE and CMAE in \cref{fig:vary_params_noise}. For clean data (first row), results align with linear algebra principles. For noisy data, we use \cref{eq:mod_gen} with additive white Gaussian noise: $\mathbf{x}^{(m)}=\mathbf{S}^{(m)}\mathbf{z}+\varepsilon, \varepsilon \sim \mathcal{N}(0,1)$ for each modality $m$. Our experiments reveal several important findings:
\begin{itemize}
    \item \textbf{Sample Size Impact:} The number of samples relative to $k$ and $d$ does not significantly affect alignment performance in noise-free conditions (Plot 1).
    \item \textbf{Dimensional Relationships:} As $d$ grows farther from $k$, CMAE decrease substantially as we have more lower-signal components (Plot 3). Similarly, as $k$ approaches $d$, error increases exponentially (Plot 5).
    \item \textbf{Optimal Configuration:} When $k < 0.5d =d_1=d_2$, CMAE approaches zero, suggesting this ratio may balance representation informativeness and alignment strength (Plots 3 and 5). This supports Remark \ref{remark:kd}.
    \item \textbf{Noise Effects Number of Samples:} In plot 7, with noise, error increases dramatically when $n > d = k$-precisely when we transition from perfect to approximate alignment (tall to wide matrix). This indicates that approximate alignment is particularly vulnerable to noise, with errors increasing by 13 orders of magnitude with just unit variance Gaussian noise. Plots 8-12 have trends that match their noiseless counterparts above them.
    \item \textbf{Latent Dimension Impact on Reconstruction}: While MLRE shows no clear pattern for $n$ and $d$ variations (plots 2,4,8,10), it consistently increases with larger $k$ values in plots 6 and 12, demonstrating the curse of dimensionality where reconstruction becomes harder in higher-dimensional spaces.
\end{itemize}

\paragraph{Limitations.}
While our method demonstrates strong performance on synthetic data, several challenges remain for real-world applications. One fundamental limitation is our assumption of linear transformations in \cref{eq:stacked_matrices}, which likely does not hold for high-dimensional raw data like image pixels and inertial signals.
Computational cost presents another challenge: perfect alignment in high-dimensional spaces is expensive, and real-world data often exhibits highly non-linear relationships, complicating direct application.
To address these issues, we propose applying our perfect alignment method to the output features of pretrained modality-specific encoders, leveraging their ability to extract meaningful representations before performing alignment. We explore this strategy using pretrained CLIP encoders as the foundation for our alignment process.

\begin{figure}
    \centering
    \includegraphics[width=.9\linewidth]{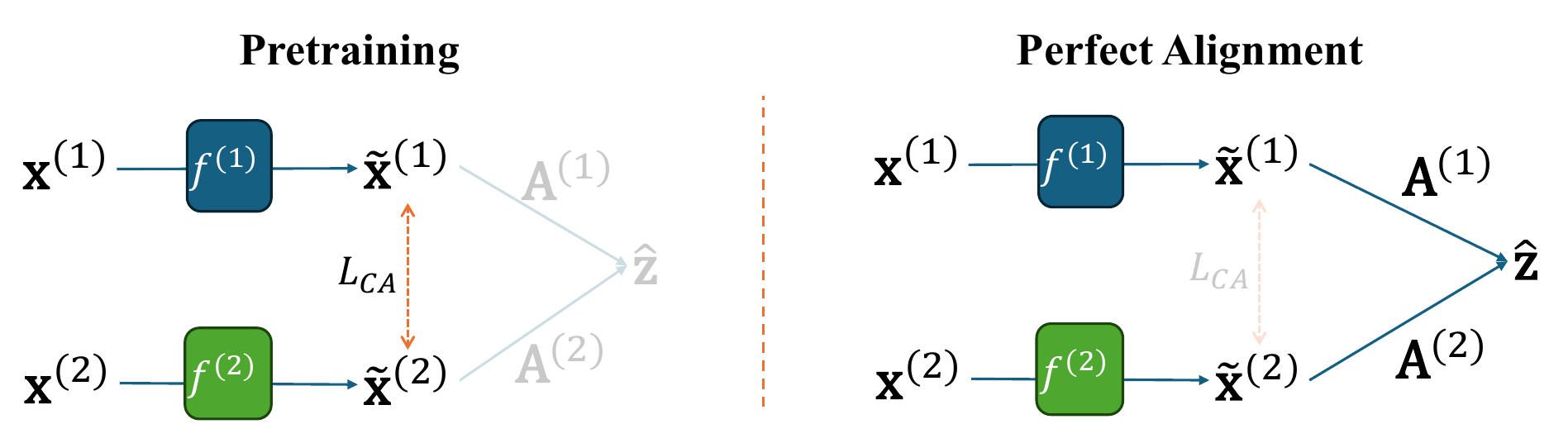}
    \caption{\textbf{CA+PA Diagram:} Perfect alignment might be applied on top of pretrained encoders. This leverages the nonlinear capabilities of deep learning as well as the perfect linear translation capabilities of our proposed perfect alignment method.}
    \label{fig:CA+PA_diagram}
\end{figure}



\subsection{Cross-modal Transfer Experiments:}
Sensor-based human action recognition (HAR) has applications in healthcare monitoring, XR interaction, and smart devices. 
However, current methods lack the generalization of video-based HAR which can leverage internet-scale datasets. 
One method to address this is by aligning videos and sensor data in a shared latent space for cross-modal transfer. 
We reproduce a contrastive alignment (CA) based cross-modal transfer method between RGB videos and Inertial Measurement Unit (IMU) data\cite{kamboj2024c3t,moon2022imu2clip,girdhar2023imagebind,gong2023mmg}, and demonstrate that our Perfect Alignment method improves HAR task accuracy.

\textbf{UTD-MHAD Dataset:} We used the UTD-Multi-modal Human Action Dataset containing 861 sequences of synchronized RGB videos and IMU data across 27 action classes performed by 8 subjects. The IMU sensor provided 3-axis acceleration and 3-axis gyroscopic information. We use RGB as our transfer source data $\mathbf{x}^{(1)} \in \mathcal{X}^{(1)}$ and IMU as our transfer target $\mathbf{x}^{(2)} \in \mathcal{X}^{(2)}$.

\textbf{Methods:}
Following \citet{kamboj2024c3t} we split the dataset into 4 disjoint subsets utilizing a 40-40-10-10 percent data split:
\begin{align*}
    \mathcal{D}_{\text{HAR}} = {\{(\mathbf{x}_i^{(1)},\mathbf{y}_i)}\}_{i=1}^{336}, 
    \mathcal{D}_{\text{Align}} = {\{(\mathbf{x}_i^{(1)},\mathbf{x}_i^{(2)}})\}_{i=1}^{336}, \\
    \mathcal{D}_{\text{Val}} = {\{(\mathbf{x}_i^{(2)},\mathbf{y}_i)}\}_{i=1}^{84},
    \mathcal{D}_{\text{Test}} = {\{(\mathbf{x}_i^{(2)},\mathbf{y}_i)}\}_{i=1}^{84}
\end{align*}
where $y \in \mathcal{Y}$ are the labels. 
We use 3 neural network modules:
\begin{enumerate}
    \item \emph{Video Feature Encoder $f^{(1)}$}: Pretrained ResNet18 followed by a 3D CNN and an 2-layer MLP
    \item \emph{IMU Feature Encoder $f^{(2)}$}: 1D CNN followed by an MLP
    \item \emph{HAR Task Decoder $h$}: MLP classifier
\end{enumerate}



For our baseline (\textbf{CA}), we perform contrastive alignment on  $\mathcal{D}_{\text{Align}}$ to learn encoders that transform raw data into correlated spaces, then train the decoder on labeled RGB data ($\mathcal{D}_{\text{HAR}}$). Finally, we test on IMU data with the IMU encoder and HAR decoder ($\mathcal{D}_{\text{Val}}$ during training evaluation and $\mathcal{D}_{\text{Test}}$ during testing). 
For our method (\textbf{CA+PA}), we add perfect alignment to recover linear transformations $\mathbf{A}^{(1)}$ and $\mathbf{A}^{(2)}$ after contrastive learning (\cref{fig:CA+PA_diagram}).

\textbf{Quantitative Results:} Table 1 demonstrates that the Perfect Alignment framework significantly improves CMAE and enhances performance in this challenging cross-modal transfer task. The CMAE decreases from 2210 to 8.66, while Top-1 accuracy increases by 5.3 percentage points. All metrics except Top-3 accuracy show improvements of at least one standard deviation, establishing statistical significance.

Unlike the synthetic experiments, zero CMAE is likely not achieved because a) real-world data is imperfect and noisy, and b) evaluation occurs on different data from the training data. The method guarantees perfect linear alignment on the training set's CA representations, but not on unseen test data. Nevertheless, the average alignment error for 2048-dimensional vectors remains remarkably low at 8.6.

\begin{table}[t]
\centering
\caption{Comparison of contrastive alignment (CA) and our method (CA+PA) on the UTD-MHAD dataset for RGB$\rightarrow$IMU transfer. Results show average $\pm$ std over 3 trials.}
\label{tab:utd_CA_PA_comparison}
\begin{tabular}{lccccc}
\toprule
\textbf{Method} & \textbf{CMAE} $\downarrow$ & \textbf{Top-1 (\%)} $\uparrow$ & \textbf{Top-3 (\%)} $\uparrow$ & \textbf{Top-5 (\%)} $\uparrow$ & \textbf{Top-7 (\%)} $\uparrow$ \\
\midrule
CA & $2210 \pm 20.5$ & $38.6 \pm 3.01$ & $56.1 \pm 6.66$ & $66.3 \pm 5.61$ & $73.9 \pm 5.21$ \\
CA+PA & $8.66 \pm 0.306$ & $43.9 \pm 4.28$ & $60.6 \pm 5.67$ & $69.7 \pm 2.83$ & $75.8 \pm 1.07$ \\
\midrule
$\Delta$ with PA& $-2200$ & $+5.30$ & $+4.50$ & $+3.40$ & $+1.90$ \\
\bottomrule
\end{tabular}
\end{table}

 \definecolor{green_1}{RGB}{44,160,44}
 \definecolor{purple}{RGB}{148,103,189}
 \definecolor{orange_1}{RGB}{255,127,14}
 \definecolor{light_green}{RGB}{188,189,34}
 \definecolor{red_1}{RGB}{214,39,40}

\begin{figure}[t]
\centering
\includegraphics[width=\linewidth]{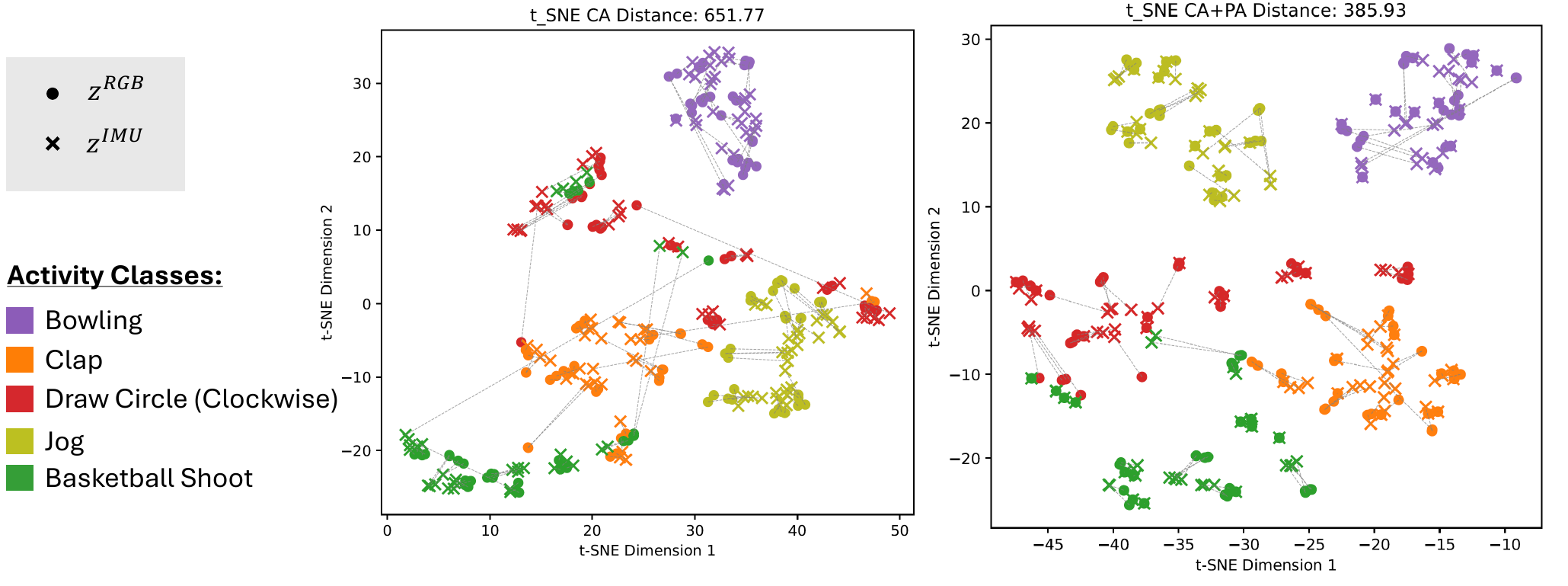}
\caption{\textbf{Contrastive Alignment TSNE Plots:} t-SNE visualization of latent representations for five activity classes after contrastive alignment of video and IMU data (CA) and after contrastive alignment and perfect alignment (CA+PA). The latent representations of the RGB (o) and IMU (x) overlap more frequently in CA+PA. We also report the total distance between each paired modality in the t-SNE space and show that CA+PA is almost half as much as CA alone, indicating PA adds a strong sense of alignment to the data.}
\label{fig:utd_latent}
\end{figure}
\textbf{Qualitative Results:}
\cref{fig:utd_latent} shows t-SNE visualizations\cite{van2008visualizing} of latent representations after CA and CA+PA. 
The RGB (o) and IMU (x) representations overlap more frequently with CA+PA, and the total distance between paired modalities (displayed in the title) is reduced by almost half, confirming that perfect alignment substantially improves modality correspondence.

\section{Related Works}

\textbf{Foundations of Contrastive Multimodal Alignment}
Alignment refers to the process of establishing correspondences between elements of different modalities, which is crucial given that different modalities have different representational structures, dimensionalities and temporal/spatial characteristics. Alignment can occur explicitly through predefined correspondences, such as cross-correlation analysis, or implicitly through training objectives, such as contrastive alignment \citet{li2024multimodal}.

The InfoNCE loss \citet{oord2018representation} formalized contrastive learning as a lower bound on mutual information. Another seminal work is SimCLR~\cite{chen2020simple}, which demonstrates that contrastive learning could rival supervised performance on ImageNet through three key insights: compositional data augmentations create effective positive pairs, nonlinear projection heads improve representation quality, and large batch sizes and training durations are critical. \citet{wang2020understanding} later proved that contrastive learning optimizes two properties: Alignment (similarity of positive pairs) and Uniformity (maximal spread of representations on the unit sphere). Later, \citet{tosh2021contrastive} developed a connection between contrastive learning and CCA to show that contrastive learning is optimal when two modalities or views provide redundant information. \citet{zheng2023contrastive} introduced Contrastive Difference Predictive Coding, combining temporal difference learning with contrastive objectives to improve sample efficiency in sequential decision-making tasks.

\textbf{Canonical Correlation Analysis and Neural Extensions}
Traditional correlation-based methods provide foundational mathematical frameworks for alignment. Hotelling's Canonical Correlation Analysis (CCA) (1936) pioneered maximizing correlation between linear projections of two views of a data, which may be seen as two modalities \citet{hotelling1992relations}. Modern extensions like Kernel CCA (KCCA) \citet{hardoon2004canonical} and Deep CCA (DCCA) \citet{andrew2013deep} learn nonlinear mappings while preserving correlation maximization principles. 
\citet{gao2017discriminative} introduced Discriminative Multiple Canonical Correlation Analysis (DMCCA) to enhance multimodal feature fusion through discriminant-aware correlation maximization, establishing a unified framework where CCA, Multiple CCA, and Discriminative CCA are special cases. \citet{li2018survey} provides a more in-depth survey on multi-view methods like CCA. 
Despite their success, these CCA methods focus on correlation rather than identity in the latent space, creating inherent limitations for explicit cross-modal transfer.

\textbf{Neural Architectures and Linear Algebraic Approaches}
\citet{theodoridis2020cross} developed a variational autoencoder framework that aligns cross-modal latent distributions via intermediate probabilistic mappings, acting as a translation mechanism between modality-specific VAEs. \citet{poklukar2022geometric} proposed Geometric Multimodal Contrastive Learning (GMC) to enforce geometric consistency across modalities through a two-level architecture with modality-specific base encoders and a shared projection head, achieving state-of-the-art performance even with missing modality information. \citet{choi2023cross} demonstrated that linear algebraic methods like SVD can achieve competitive cross-modal retrieval without neural training, using a simple mapping computed from least squares and singular value decomposition, however, their method maps one modality directly to the other, whereas our framework provides two transformations, one for each modality.

The proposed perfect alignment method bridges these approaches by formulating alignment as a linear inverse problem solvable via SVD, connecting the theoretical rigor of linear algebraic methods with the practical success of neural alignment techniques.

\section{Conclusion}
\label{sec:conclusion}
\textbf{Limitations:}
This work focuses on aligning two modalities and tests on RGB to IMU transfer; however, extending to more than two modalities presents additional challenges. The application is limited to human activity recognition, though testing perfect alignment across multiple tasks would provide valuable insights. Different task requirements may benefit differently from the same alignment between modalities, as certain joint representations may be more informative for specific tasks.

The method is unlikely to perform well with textual data due to the inherent many-to-many mapping between text and other modalities, precluding true perfect alignment. For instance, a single image of a dog corresponds to numerous possible textual descriptions, and vice versa. In such cases, learned approximate objectives or correlation-based approaches where representations are similar but not identical may be more appropriate.

Additionally, the current approach is constrained by its linear mapping assumption when stacking transformations $\mathbf{A}^{(1)}$ and $\mathbf{A}^{(2)}$. Future work could explore non-linear mappings, potentially through compositions of transformations or iterative constructions based on SVD results and linear transformations.

\paragraph{Summary.}
This paper introduced a method for perfect multimodal alignment by formulating the problem as an inverse projection onto a shared latent space. Experiments on synthetic data demonstrate near-zero alignment error. Importantly, the transformation preserves the relative structure of the data, potentially allowing for class identification in the estimated latent space. These results highlight the promise of our alignment technique for multimodal analysis and motivate further research into perfect alignment on complex, real-world datasets.

\begin{ack}
This material is based upon work supported by the National Science Foundation (NSF) Graduate Research Fellowship Program.
We acknowledge the NSF for providing access to the NCSA Delta GPU cluster (A100 and A40 GPUs) under the NSF ACCESS program (allocation project CIS240405). This resource enabled the computational experiments critical to this research.
\end{ack}

\bibliographystyle{plainnat}
\bibliography{gaussian_align}

\begin{thebibliography}{27}
\providecommand{\natexlab}[1]{#1}
\providecommand{\url}[1]{\texttt{#1}}
\expandafter\ifx\csname urlstyle\endcsname\relax
  \providecommand{\doi}[1]{doi: #1}\else
  \providecommand{\doi}{doi: \begingroup \urlstyle{rm}\Url}\fi

\bibitem[Andrew et~al.(2013)Andrew, Arora, Bilmes, and Livescu]{andrew2013deep}
Galen Andrew, Raman Arora, Jeff Bilmes, and Karen Livescu.
\newblock Deep canonical correlation analysis.
\newblock In \emph{International conference on machine learning}, pages 1247--1255. PMLR, 2013.

\bibitem[Che and Eysenbach(2025)]{che2025law}
Yongwei Che and Benjamin Eysenbach.
\newblock The" law" of the unconscious contrastive learner: Probabilistic alignment of unpaired modalities.
\newblock \emph{arXiv preprint arXiv:2501.11326}, 2025.

\bibitem[Chen et~al.(2020)Chen, Kornblith, Norouzi, and Hinton]{chen2020simple}
Ting Chen, Simon Kornblith, Mohammad Norouzi, and Geoffrey Hinton.
\newblock A simple framework for contrastive learning of visual representations.
\newblock In \emph{International conference on machine learning}, pages 1597--1607. PmLR, 2020.

\bibitem[Chen et~al.(2024)Chen, Lin, Liu, Xiao, and Dyer]{chen2024your}
Zihao Chen, Chi-Heng Lin, Ran Liu, Jingyun Xiao, and Eva Dyer.
\newblock Your contrastive learning problem is secretly a distribution alignment problem.
\newblock \emph{Advances in Neural Information Processing Systems}, 37:\penalty0 91597--91617, 2024.

\bibitem[Choi et~al.(2023)Choi, Lee, Joe, and Gwon]{choi2023cross}
Hyunjin Choi, Hyunjae Lee, Seongho Joe, and Youngjune Gwon.
\newblock Is cross-modal information retrieval possible without training?
\newblock In \emph{European Conference on Information Retrieval}, pages 377--385. Springer, 2023.

\bibitem[Gao et~al.(2017)Gao, Qi, Chen, and Guan]{gao2017discriminative}
Lei Gao, Lin Qi, Enqing Chen, and Ling Guan.
\newblock Discriminative multiple canonical correlation analysis for information fusion.
\newblock \emph{IEEE Transactions on Image Processing}, 27\penalty0 (4):\penalty0 1951--1965, 2017.

\bibitem[Girdhar et~al.(2023{\natexlab{a}})Girdhar, El-Nouby, Liu, Singh, Alwala, Joulin, and Misra]{Girdhar_2023_CVPR}
Rohit Girdhar, Alaaeldin El-Nouby, Zhuang Liu, Mannat Singh, Kalyan~Vasudev Alwala, Armand Joulin, and Ishan Misra.
\newblock Imagebind: One embedding space to bind them all.
\newblock In \emph{Proceedings of the IEEE/CVF Conference on Computer Vision and Pattern Recognition (CVPR)}, pages 15180--15190, June 2023{\natexlab{a}}.

\bibitem[Girdhar et~al.(2023{\natexlab{b}})Girdhar, El-Nouby, Liu, Singh, Alwala, Joulin, and Misra]{girdhar2023imagebind}
Rohit Girdhar, Alaaeldin El-Nouby, Zhuang Liu, Mannat Singh, Kalyan~Vasudev Alwala, Armand Joulin, and Ishan Misra.
\newblock Imagebind: One embedding space to bind them all.
\newblock In \emph{Proceedings of the IEEE/CVF Conference on Computer Vision and Pattern Recognition}, pages 15180--15190, 2023{\natexlab{b}}.

\bibitem[Gong et~al.(2023)Gong, Mohan, Dhingra, Bazin, Li, Wang, and Ranjan]{gong2023mmg}
Xinyu Gong, Sreyas Mohan, Naina Dhingra, Jean-Charles Bazin, Yilei Li, Zhangyang Wang, and Rakesh Ranjan.
\newblock Mmg-ego4d: Multimodal generalization in egocentric action recognition.
\newblock In \emph{Proceedings of the IEEE/CVF Conference on Computer Vision and Pattern Recognition}, pages 6481--6491, 2023.

\bibitem[Hardoon et~al.(2004)Hardoon, Szedmak, and Shawe-Taylor]{hardoon2004canonical}
David~R Hardoon, Sandor Szedmak, and John Shawe-Taylor.
\newblock Canonical correlation analysis: An overview with application to learning methods.
\newblock \emph{Neural computation}, 16\penalty0 (12):\penalty0 2639--2664, 2004.

\bibitem[Harikumar and Bresler(1999)]{harikumar1999perfect}
Gopal Harikumar and Yoram Bresler.
\newblock Perfect blind restoration of images blurred by multiple filters: Theory and efficient algorithms.
\newblock \emph{IEEE Transactions on Image Processing}, 8\penalty0 (2):\penalty0 202--219, 1999.

\bibitem[Hotelling(1992)]{hotelling1992relations}
Harold Hotelling.
\newblock Relations between two sets of variates.
\newblock In \emph{Breakthroughs in statistics: methodology and distribution}, pages 162--190. Springer, 1992.

\bibitem[Huh et~al.(2024)Huh, Cheung, Wang, and Isola]{huh2024platonic}
Minyoung Huh, Brian Cheung, Tongzhou Wang, and Phillip Isola.
\newblock Position: The platonic representation hypothesis.
\newblock In \emph{Proceedings of the 41st International Conference on Machine Learning}, volume 235 of \emph{Proceedings of Machine Learning Research}, pages 20617--20642. PMLR, 2024.
\newblock URL \url{https://proceedings.mlr.press/v235/huh24a.html}.

\bibitem[Kamboj et~al.(2024)Kamboj, Nguyen, and Do]{kamboj2024c3t}
Abhi Kamboj, Anh~Duy Nguyen, and Minh Do.
\newblock C3t: Cross-modal transfer through time for human action recognition.
\newblock \emph{arXiv preprint arXiv:2407.16803}, 2024.

\bibitem[Li and Tang(2024)]{li2024multimodal}
Songtao Li and Hao Tang.
\newblock Multimodal alignment and fusion: A survey.
\newblock \emph{arXiv preprint arXiv:2411.17040}, 2024.

\bibitem[Li et~al.(2018)Li, Yang, and Zhang]{li2018survey}
Yingming Li, Ming Yang, and Zhongfei Zhang.
\newblock A survey of multi-view representation learning.
\newblock \emph{IEEE transactions on knowledge and data engineering}, 31\penalty0 (10):\penalty0 1863--1883, 2018.

\bibitem[Moon et~al.(2022)Moon, Madotto, Lin, Dirafzoon, Saraf, Bearman, and Damavandi]{moon2022imu2clip}
Seungwhan Moon, Andrea Madotto, Zhaojiang Lin, Alireza Dirafzoon, Aparajita Saraf, Amy Bearman, and Babak Damavandi.
\newblock Imu2clip: Multimodal contrastive learning for imu motion sensors from egocentric videos and text.
\newblock \emph{arXiv preprint arXiv:2210.14395}, 2022.

\bibitem[Nakada et~al.(2023)Nakada, Gulluk, Deng, Ji, Zou, and Zhang]{nakada2023understanding}
Ryumei Nakada, Halil~Ibrahim Gulluk, Zhun Deng, Wenlong Ji, James Zou, and Linjun Zhang.
\newblock Understanding multimodal contrastive learning and incorporating unpaired data.
\newblock In \emph{International Conference on Artificial Intelligence and Statistics}, pages 4348--4380. PMLR, 2023.

\bibitem[Oord et~al.(2018)Oord, Li, and Vinyals]{oord2018representation}
Aaron van~den Oord, Yazhe Li, and Oriol Vinyals.
\newblock Representation learning with contrastive predictive coding.
\newblock \emph{arXiv preprint arXiv:1807.03748}, 2018.

\bibitem[Poklukar et~al.(2022)Poklukar, Vasco, Yin, Melo, Paiva, and Kragic]{poklukar2022geometric}
Petra Poklukar, Miguel Vasco, Hang Yin, Francisco~S Melo, Ana Paiva, and Danica Kragic.
\newblock Geometric multimodal contrastive representation learning.
\newblock In \emph{International Conference on Machine Learning}, pages 17782--17800. PMLR, 2022.

\bibitem[Poole et~al.(2019)Poole, Ozair, Van Den~Oord, Alemi, and Tucker]{poole2019variational}
Ben Poole, Sherjil Ozair, Aaron Van Den~Oord, Alex Alemi, and George Tucker.
\newblock On variational bounds of mutual information.
\newblock In \emph{International conference on machine learning}, pages 5171--5180. PMLR, 2019.

\bibitem[Radford et~al.(2021)Radford, Kim, Hallacy, Ramesh, Goh, Agarwal, Sastry, Askell, Mishkin, Clark, Krueger, and Sutskever]{radford2021learning}
Alec Radford, Jong~Wook Kim, Chris Hallacy, Aditya Ramesh, Gabriel Goh, Sandhini Agarwal, Girish Sastry, Amanda Askell, Pamela Mishkin, Jack Clark, Gretchen Krueger, and Ilya Sutskever.
\newblock Learning transferable visual models from natural language supervision.
\newblock In Marina Meila and Tong Zhang, editors, \emph{Proceedings of the 38th International Conference on Machine Learning}, volume 139 of \emph{Proceedings of Machine Learning Research}, pages 8748--8763. PMLR, 18--24 Jul 2021.
\newblock URL \url{https://proceedings.mlr.press/v139/radford21a.html}.

\bibitem[Theodoridis et~al.(2020)Theodoridis, Chatzis, Solachidis, Dimitropoulos, and Daras]{theodoridis2020cross}
Thomas Theodoridis, Theocharis Chatzis, Vassilios Solachidis, Kosmas Dimitropoulos, and Petros Daras.
\newblock Cross-modal variational alignment of latent spaces.
\newblock In \emph{Proceedings of the IEEE/CVF conference on computer vision and pattern recognition workshops}, pages 960--961, 2020.

\bibitem[Tosh et~al.(2021)Tosh, Krishnamurthy, and Hsu]{tosh2021contrastive}
Christopher Tosh, Akshay Krishnamurthy, and Daniel Hsu.
\newblock Contrastive learning, multi-view redundancy, and linear models.
\newblock In \emph{Algorithmic Learning Theory}, pages 1179--1206. PMLR, 2021.

\bibitem[Van~der Maaten and Hinton(2008)]{van2008visualizing}
Laurens Van~der Maaten and Geoffrey Hinton.
\newblock Visualizing data using t-sne.
\newblock \emph{Journal of machine learning research}, 9\penalty0 (11), 2008.

\bibitem[Wang and Isola(2020)]{wang2020understanding}
Tongzhou Wang and Phillip Isola.
\newblock Understanding contrastive representation learning through alignment and uniformity on the hypersphere.
\newblock In \emph{International conference on machine learning}, pages 9929--9939. PMLR, 2020.

\bibitem[Zheng et~al.(2023)Zheng, Salakhutdinov, and Eysenbach]{zheng2023contrastive}
Chongyi Zheng, Ruslan Salakhutdinov, and Benjamin Eysenbach.
\newblock Contrastive difference predictive coding.
\newblock \emph{arXiv preprint arXiv:2310.20141}, 2023.

\end{thebibliography}








\newpage
\section{Appendix}

\subsection{Discussion on Using Text Labels as a Modality}
\label{app:discussion}
Our method exploits a one-to-one correspondence between modality spaces $\mathcal{X}^{1}$ and $\mathcal{X}^{2}$. When a single instance in modality 1 corresponds to multiple instances in modality 2, the data vectors in $\mathbf{X}^{(1)}$ exhibit redundancy, resulting in a stacked matrix $\mathbf{X}$ with diminished operator norm. This phenomenon was empirically observed in experiments with CIFAR-10, where we utilized class labels as the first modality and images as the second modality, resulting in alignment matrices with values approaching zero.

Textual data presents particular challenges for perfect alignment due to its inherent semantic ambiguity-the term "dog" encompasses numerous breeds with distinct visual characteristics. Conversely, two canine images will invariably exhibit unique pixel configurations. Similar distinctions apply to various sensor modalities, including depth maps, LiDAR scans, and inertial measurement unit (IMU) data.




\subsection{Proofs of Theorem \ref{thm:perfect_alignment} and Corollary \ref{cor:approximate_alignment}}
\label{app:proof_perfect_alignment}

\textbf{Theorem \ref{thm:perfect_alignment} Restated} (Existence of Perfect Alignment).
Given the inverse problem $\mathbf{A}\mathbf{X} = \mathbf{0}$ defined in \cref{eq:perf_align}, where $\mathbf{X} \in \mathbb{R}^{d \times n}$ is a given data matrix and $\mathbf{A} \in \mathbb{R}^{k \times d}$ is unknown, if $\mathbf{X}$ has a left null space $\mathcal{N}(\mathbf{X}^T)$ of dimension $\dim(\mathcal{N}(\mathbf{X}^T)) \geq k$, then there exists a closed-form solution for $\mathbf{A}$. Specifically, the rows of $\mathbf{A}$ can be formed by any $k$ linearly independent vectors spanning $\mathcal{N}(\mathbf{X}^T)$.

\begin{proof}
Let $\mathcal{N}(\mathbf{X}^T)$ denote the left null space of $\mathbf{X}$, defined as:
$$
\mathcal{N}(\mathbf{X}^T) = \{ \mathbf{y} \in \mathbb{R}^d \mid \mathbf{X}^T\mathbf{y} = \mathbf{0} \}
$$
By the rank-nullity theorem:
$$
\dim(\mathcal{N}(\mathbf{X}^T)) = d - \rank(\mathbf{X})
$$
The theorem assumes $\dim(\mathcal{N}(\mathbf{X}^T)) \geq k$. Therefore, there exist $k$ linearly independent vectors $\{\mathbf{v}_1, \mathbf{v}_2, \dots, \mathbf{v}_k\} \subseteq \mathcal{N}(\mathbf{X}^T)$.

Construct $\mathbf{A} \in \mathbb{R}^{k \times d}$ by setting these vectors as its rows:
\[
\mathbf{A} = \begin{bmatrix}
\mathbf{v}_1^T \\
\mathbf{v}_2^T \\
\vdots \\
\mathbf{v}_k^T
\end{bmatrix}
\]

For each row $\mathbf{v}_i^T$ of $\mathbf{A}$, we have:
\[
\mathbf{v}_i^T \mathbf{X} = \mathbf{0}^T \quad \text{(since } \mathbf{v}_i \in \mathcal{N}(\mathbf{X}^T)\text{)}
\]
Therefore, the matrix product satisfies:
\[
\mathbf{A}\mathbf{X} = \begin{bmatrix}
\mathbf{v}_1^T \mathbf{X} \\
\mathbf{v}_2^T \mathbf{X} \\
\vdots \\
\mathbf{v}_k^T \mathbf{X}
\end{bmatrix} = \begin{bmatrix}
\mathbf{0}^T \\
\mathbf{0}^T \\
\vdots \\
\mathbf{0}^T
\end{bmatrix} = \mathbf{0}
\]

The rows of $\mathbf{A}$ are linearly independent by construction, as they form a basis for a $k$-dimensional subspace of $\mathcal{N}(\mathbf{X}^T)$. This completes the proof that such an $\mathbf{A}$ exists and satisfies $\mathbf{A}\mathbf{X} = \mathbf{0}$.

The closed-form solution arises from the fact that $\mathcal{N}(\mathbf{X}^T)$ can be explicitly computed via:
\begin{itemize}
\item Singular Value Decomposition (SVD): If $\mathbf{X} = \mathbf{U}\mathbf{\Sigma} \mathbf{V}^T$, then $\mathcal{N}(\mathbf{X}^T)$ is spanned by the last $d - \rank(\mathbf{X})$ columns of $\mathbf{U}$.
\item Reduced Row Echelon Form: For $\mathbf{X}^T$, the null space basis vectors correspond to the free variables in $\mathrm{rref}(\mathbf{X}^T)$.
\end{itemize}

Thus, any $k$ linearly independent vectors from these computed bases will satisfy the requirements for $\mathbf{A}$.
\end{proof}

\textbf{Corollary \ref{cor:approximate_alignment} Restated} (Approximate Alignment).
\textit{If $\mathbf{X} \in \mathbb{R}^{d \times n}$ has a left null space $\mathcal{N}(\mathbf{X}^T)$ with $\dim(\mathcal{N}(\mathbf{X}^T)) < k$, an approximation to $\mathbf{A}\mathbf{X} = \mathbf{0}$ can be obtained by selecting the $k$ basis vectors corresponding to the smallest singular values of $\mathbf{X}$. This approximation minimizes the Frobenius norm $\|\mathbf{A}\mathbf{X}\|_F$.}

\begin{proof}
Let $\mathbf{X} = \mathbf{U}\mathbf{\Sigma} \mathbf{V}^T$ be the SVD of $\mathbf{X}$, where $\mathbf{U} \in \mathbb{R}^{d \times d}$ and $\mathbf{V} \in \mathbb{R}^{n \times n}$ are orthogonal matrices, and $\mathbf{\Sigma} \in \mathbb{R}^{d \times n}$ contains the singular values $\sigma_1 \geq \sigma_2 \geq \dots \geq \sigma_r \geq 0$ with $r = \rank(\mathbf{X})$. 

The left null space of $\mathbf{X}$ is spanned by columns of $\mathbf{U}$ corresponding to zero singular values. When $\dim(\mathcal{N}(\mathbf{X}^T)) < k$, we instead use the $k$ columns of $\mathbf{U}$ associated with the smallest singular values $\sigma_{d - k + 1}, \dots, \sigma_d$. Constructing $\mathbf{A}$ as:
\[
\mathbf{A} = \begin{bmatrix} \mathbf{u}_{d - k + 1}^T \\ \vdots \\ \mathbf{u}_d^T \end{bmatrix},
\]
where $\mathbf{u}_i$ is the $i^{\text{th}}$ column of $\mathbf{U}$. Letting $\mathbf{v}_i^T$ representing the $i^{\text{th}}$ row of $\mathbf{V}^T$ we compute:
\[
\mathbf{A}\mathbf{X} = \begin{bmatrix} \sigma_{d - k + 1} \mathbf{v}_{d - k + 1}^T \\ \vdots \\ \sigma_d \mathbf{v}_d^T \end{bmatrix}.
\]
The Frobenius norm $\|\mathbf{A}\mathbf{X}\|_F^2 = \sum_{i=d - k + 1}^d \sigma_i^2$ is minimized because the Eckart-Young-Mirsky theorem ensures that truncating to the smallest $k$ singular values yields the optimal low-rank approximation in the Frobenius norm. Any other choice of vectors would include larger singular values, increasing the norm.
\end{proof}

\subsection{Additional Experiments:}
\begin{table}[h]
\centering
\resizebox{.4\linewidth}{!}{
\begin{tabular}{lc}
\toprule
\textbf{Error Metric} & \textbf{Value} \\
\midrule
MRE using ${\mathbf{S}^{(1)}}^\dagger$ & $2.98 \times 10^{-16}$ \\
MRE using ${\mathbf{S}^{(2)}}^\dagger$ & $6.47 \times 10^{-16}$ \\
\bottomrule
\end{tabular}
}
\caption{Sanity check: MLRE when using pseudo-inverse of ground-truth transformation matrices ${\mathbf{S}^{(m)}}^\dagger$. These near-zero errors validate our MLRE metric's ability to detect perfect reconstruction.}
\label{tab:synth_errors_sanity}
\end{table}

\subsection{Notation Reference:}
We construct a table of our notation in \cref{tab:notation_reference}
\begin{table*}[t]
\centering
\caption{Notation Reference}
\label{tab:notation_reference}
\begin{tabular}{@{}lll@{}}
\toprule
\textbf{Symbol} & \textbf{Description} & \textbf{Domain/Type} \\
\midrule
$\mathcal{X}^{(m)}$ & Input space of modality $m$ & Vector space \\
$\mathcal{Z}$ & Shared latent space & Vector space \\
$\mathbf{x}_i^{(m)}$ & Data point $i$ from modality $m$ & $\mathbb{R}^{d_m}$ (column vector) \\
$\mathbf{S}^{(m)}$ & Ground-truth transformation matrix for modality $m$ & $\mathbb{R}^{d_m \times k}$ \\
$\mathbf{z}_i$ & Latent concept vector for data point $i$ & $\mathbb{R}^{k}$ \\
$f^{(m)}$ & Encoder function for modality $m$ & $\mathcal{X}^{(m)} \to \mathcal{Z}$ \\
$\mathbf{A}^{(m)}$ & Learned projection matrix for modality $m$ & $\mathbb{R}^{k \times d_m}$ \\
$\mathbf{X}^{(m)}$ & Data matrix for modality $m$ & $\mathbb{R}^{d_m \times n}$ \\
$\mathbf{A}$ & Combined projection matrix & $\mathbb{R}^{k \times d}$ ($d = \sum d_m$) \\
$\mathbf{X}$ & Stacked data matrix & $\mathbb{R}^{d \times n}$ \\
$\mathbf{0}$ & Zero matrix in $\mathbf{A}\mathbf{X} = \mathbf{0}$ & $\mathbb{R}^{k \times n}$ \\
$\mathbf{U},\mathbf{\Sigma},\mathbf{V}^T$ & SVD components of $\mathbf{X}$ & $\mathbf{U} \in \mathbb{R}^{d \times d}$, $\mathbf{\Sigma} \in \mathbb{R}^{d \times n}$, $\mathbf{V} \in \mathbb{R}^{n \times n}$ \\
$\pi_1, \pi_2$ & Mixture weights for GMM & $\pi_1 + \pi_2 = 1$ \\
$\boldsymbol{\mu}_1, \boldsymbol{\mu}_2$ & GMM mean vectors & $\mathbb{R}^{2}$ \\
$\boldsymbol{\Sigma}_1, \boldsymbol{\Sigma}_2$ & GMM covariance matrices & $\mathbb{R}^{2 \times 2}$ \\
$n$ & Number of data points & $\mathbb{N}$ \\
$d_m$ & Dimension of modality $m$ & $\mathbb{N}$ \\
$d$ & Combined data dimension & $d = \sum d_m$ \\
$k$ & Latent space dimension & $\mathbb{N}$ \\
$- - -$ & Matrix concatenation operator & -- \\
$\mathbf{I}_d$ & Identity matrix & $\mathbb{R}^{d \times d}$ \\
\bottomrule
\end{tabular}
\end{table*}

\subsection{Extension to linear decision boundary:}
\begin{figure}
    \begin{subfigure}[b]{0.48\textwidth}
        \centering
        \includegraphics[width=\textwidth]{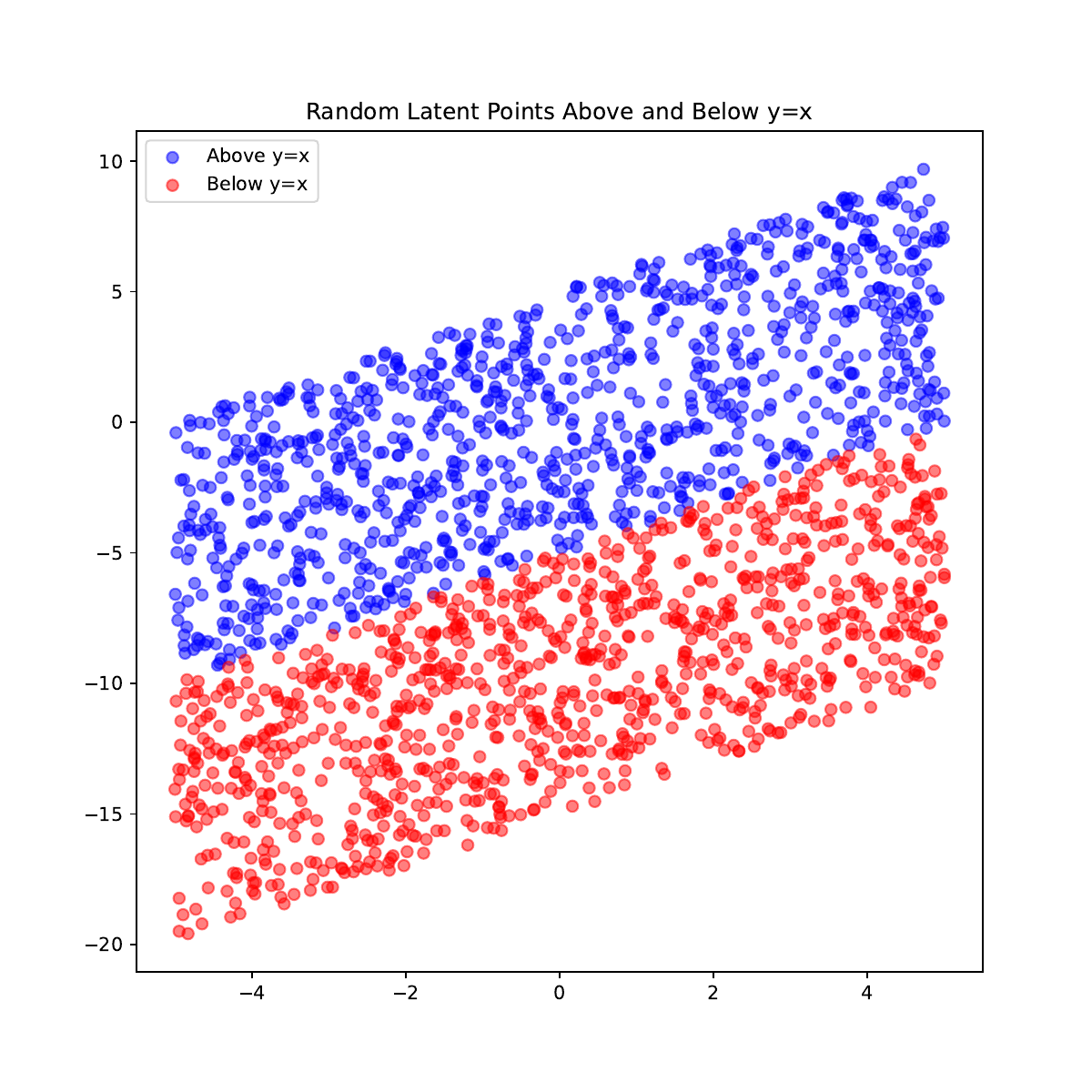}
        \caption{\textbf{Ground Truth Latent Space:} This latent space was generated by sampling 2000 random points uniformly, 1000 above and below the line y = x-10 by a maximum margin of 10. The different colors represent different classes}
        \label{fig:linear_latent}
    \end{subfigure}
    \hspace{1em}
    \begin{subfigure}[b]{0.48\textwidth}
        \centering
        \includegraphics[width=\textwidth]{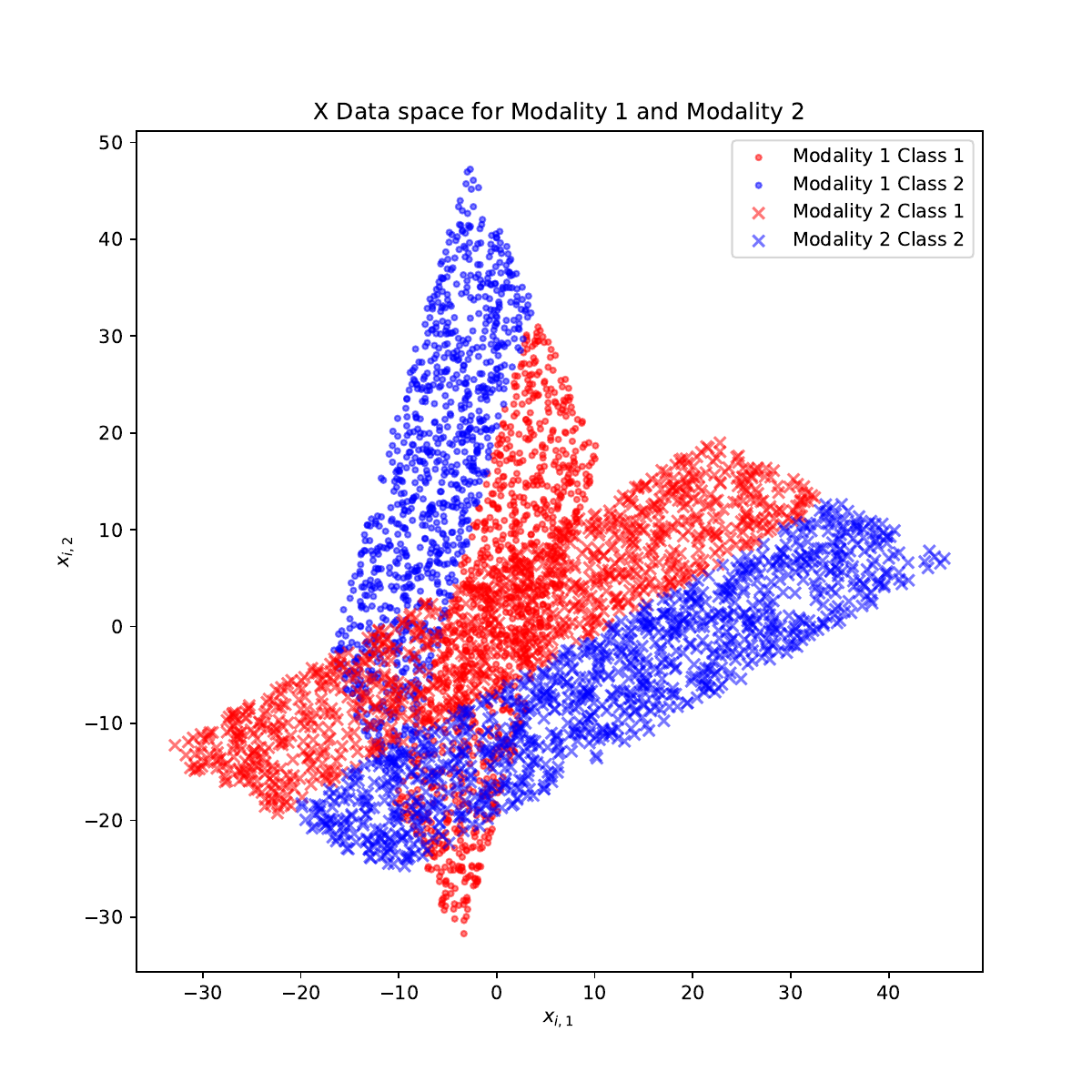}
        \caption{\textbf{Projected X Data:} This visualizes the $\mathcal{X}^{(1)}$ and $\mathcal{X}^{(2)}$ distributions generated by projecting the the points generated in \cref{fig:linear_latent}}.
        \label{fig:linear_latent_data_space}
    \end{subfigure}
    \hfill
    \begin{subfigure}[b]{.8\textwidth}
        \centering
        \includegraphics[width=\textwidth]{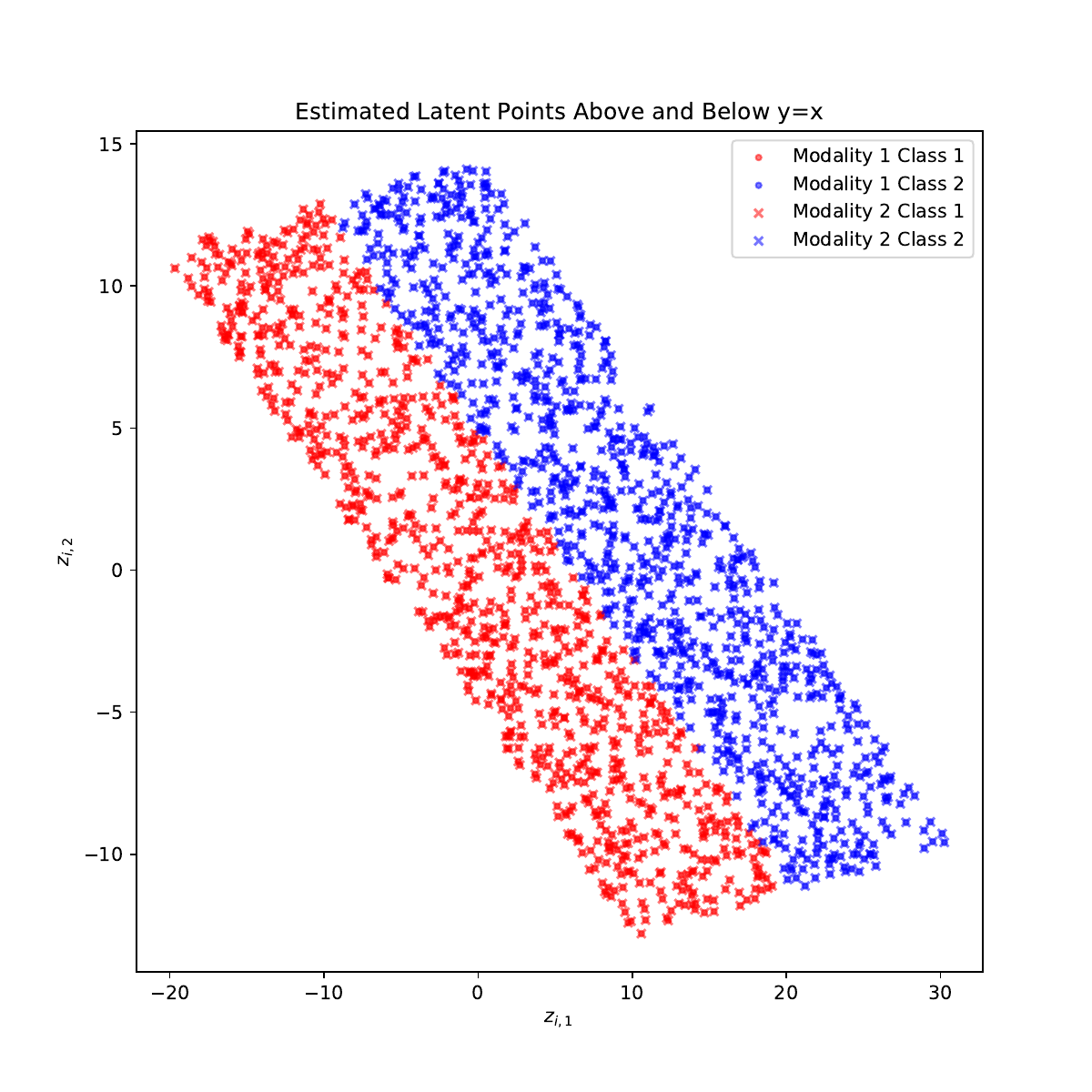}
        \caption{\textbf{Estimated Latent Space:} This is the estimated latent space generated from our perfect alignment method. Although the method didn't recover the exact latent, the structure of the latent space remains the same ,and the points from different classes are linearly separable. Furthermore, the x's and dots overlap exactly, indicating perfect alignment was achieved.}
        \label{fig:linear_estimated_Z}
    \end{subfigure}
    \caption{\textbf{Uniform Latent Concept Distributions:} These experiments demonstrate that the proposed perfect alignment cross-modal transfer method holds assuming a uniform latent concept distribution with a linear boundary.}
    \label{fig:linear_decision_boundary}
\end{figure}

Our experiments thus far have assumed the ground truth latent space $\mathcal{Z}$ contains concepts following a guassian distribution. 
We also conduct experiments assuming the underlying elements are  uniformly distributed and the concept classes are divided by a linear decision boundary.
We test this assuming two modalities, two concept classes and 2000 points, similar to our synthetic experiments in the main paper. 
The results indicate a \textbf{CMAE of 1.36 e-14} and an \textbf{MLRE of 15.6}.
See \cref{fig:linear_decision_boundary} for a visualization experiment.

\end{document}